\pdfoutput=1

\documentclass[11pt]{article}

\usepackage[final]{ACL2023}

\usepackage{times}
\usepackage{latexsym}

\usepackage[T1]{fontenc}

\usepackage[utf8]{inputenc}
\usepackage{graphicx}

\usepackage{microtype}

\usepackage{inconsolata}
\usepackage{amsmath}
\usepackage{fix-cm}

\title{Improving the Efficiency of Long Document Classification using Sentence Ranking Approach}

\author{
    Prathamesh Kokate\textsuperscript{1,3}, Mitali Sarnaik\textsuperscript{1,3}, Manavi Khopade\textsuperscript{1,3}, and Raviraj Joshi\textsuperscript{2,3} \\
    \textsuperscript{1}Pune Institute of Computer Technology, Pune \\
    \textsuperscript{2}Indian Institute of Technology Madras, Chennai \\
     \textsuperscript{3}L3Cube Labs, Pune
}

\begin{document}
\maketitle
\begin{abstract}
Long document classification poses challenges due to the computational limitations of transformer-based models, particularly BERT, which are constrained by fixed input lengths and quadratic attention complexity. Moreover, using the full document for classification is often redundant, as only a subset of sentences typically carries the necessary information. To address this, we propose a TF-IDF-based sentence ranking method that improves efficiency by selecting the most informative content. Our approach explores fixed-count and percentage-based sentence selection, along with an enhanced scoring strategy combining normalized TF-IDF scores and sentence length. Evaluated on the MahaNews LDC dataset of long Marathi news articles, the method consistently outperforms baselines such as first, last, and random sentence selection. With MahaBERT-v2, we achieve near-identical classification accuracy with just a 0.33 percent drop compared to the full-context baseline, while reducing input size by over 50 percent and inference latency by 43 percent. This demonstrates that significant context reduction is possible without sacrificing performance, making the method practical for real-world long document classification tasks.
\end{abstract}

\section{Introduction}
\begin{figure*}
  \centering
  \includegraphics[width=\textwidth]{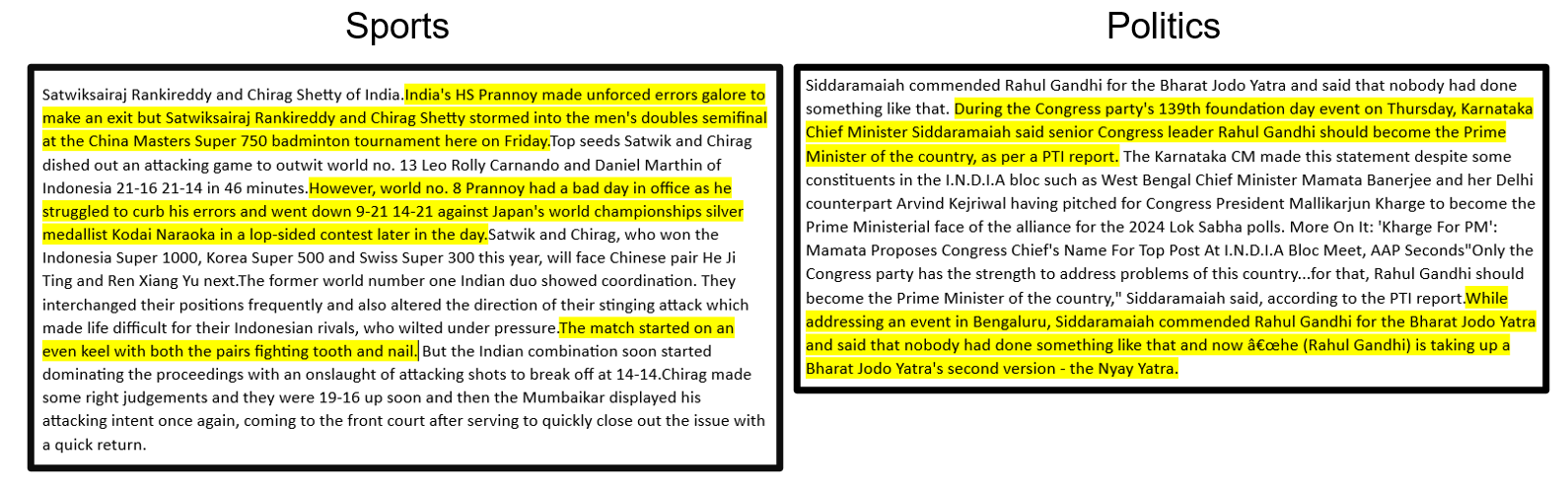}
  \caption{Illustration of key idea — selective sentence processing for efficient document classification. The figure presents two example paragraphs: one related to sports and the other to politics. In each case, the most semantically relevant and contextually informative sentences are highlighted. These highlighted sentences contain domain-specific cues (e.g., sports activities or political entities) that enable accurate classification without processing the full document. This demonstrates that selective sentence extraction can preserve classification performance while reducing computational overhead.}
  \label{fig:paragraph}
\end{figure*}

Long document classification is a fundamental task in natural language processing (NLP) with applications in categorizing research papers, legal documents, news articles, and customer reviews. Transformer-based models, such as BERT and its variants, have demonstrated state-of-the-art performance in text classification. However, these models face fundamental constraints when processing lengthy documents due to their restricted input size and exponential attention overhead (\citet{Park_2022}),(\citet{Zaheer_2020}). Standard BERT models truncate lengthy inputs, leading to information loss, or require additional mechanisms such as hierarchical processing, which increases computational overhead (\citet{Wagh_2021}). As a result, efficient handling of long texts while maintaining classification accuracy remains an open challenge (\citet{Devlin_2018}).
\newline
Traditional approaches to address this issue involve architectural modifications, such as sparse attention mechanisms or hierarchical models. However, these methods often introduce additional complexity and computational overhead. In contrast, we propose a data-driven optimization technique that reduces input size while preserving essential contextual information, allowing standard transformer models to efficiently classify long documents (\citet{Minaee_2021}). Instead of modifying the model architecture, we focus on optimizing the input representation by selecting only the most relevant sentences using a ranking mechanism.
\newline
Our approach leverages TF-IDF based sentence ranking to extract key sentences, reducing redundant information and minimizing input length (\citet{Qaiser_2018}). Sentences are ranked by computing the TF-IDF scores of individual words within each sentence, treating each sentence in a single article as a separate document (\citet{Das_2020,Kim_2019}). The total TF-IDF score for a sentence is obtained by summing the scores of its words, with the highest scoring sentence assigned the top rank (\citet{Liu_2018,Das_2023}).

Figure \ref{fig:paragraph} illustrates the key idea proposed in this work. It shows examples from sports and politics, where only the most semantically relevant sentences are highlighted. These selected sentences, rich in domain-specific terms and context, are sufficient for accurate document classification, reducing the need for full-text processing.
This highlights the efficiency of our approach in processing only a fraction of the document by selecting key sentences for classification, reducing computational overhead.
\newline
We explore multiple strategies for selecting sentences, including the following:

\begin{enumerate}
    \item Fixed-length selection involves choosing a predefined number of top-ranked sentences, with evaluations conducted for 1, 2, 3, 4, and 5 sentences.
    \item Percentage-based selection refers to the selection of a specific percentage of top-ranked sentences, varying from 10\% to 100\% in increments of 10\%.
    \item Weighted ranking combines normalized TF-IDF scores with sentence length to balance importance and informativeness, exploring different weighting factors to identify the optimal configuration.
\end{enumerate}
To evaluate the effectiveness of these strategies, we conduct extensive experiments on the MahaNews dataset (\citet{Mittal_2024,Mirashi_2024}), a corpus of long Marathi news articles categorized by topic. Using MahaBERT (marathi-bert-v2) (\citet{joshi2022l3cubeMahaCorpus}), we train and test models on reduced-context versions of the dataset and compare the classification performance across different selection methods. Our results demonstrate that TF-IDF-based ranking significantly outperforms simpler selection strategies, such as choosing the first, last, or randomly sampled sentences. Additionally, integrating length-aware weighting further enhances accuracy, while context reduction leads to a substantial decrease in inference time 
without compromising performance.
\subsection{Key Contributions:}
\begin{itemize}
    \item We propose a novel sentence ranking technique to enhance the efficiency of BERT models. This approach aims to significantly reduce the processing time required for handling large text inputs.
    \item We introduce a TF-IDF-based context reduction strategy to enhance transformer-based long document classification without modifying the model architecture.
    \item Multiple sentence selection techniques are evaluated, including fixed-length, percentage-based, and weighted ranking, highlighting their trade-offs.
    \item Experiments on the MahaNews dataset show that ranked selection consistently outperforms naive approaches while maintaining accuracy and significantly reducing inference time. Specifically, the performance of selection strategies follows the order: ranked > first > random > last. Notably, selecting sentences from the beginning of the document serves as a strong baseline.
    \item Our findings reveal an optimal balance between input length, accuracy, and computational efficiency, demonstrating that selecting a subset of ranked sentences can achieve near-full-document classification performance.
\end{itemize}
By systematically analyzing context reduction techniques, our work provides a practical and efficient alternative to architectural modifications for long document classification in transformer-based models.

\section{Related Work}

Due to the vast amount of information contained in long documents, directly processing them with traditional classification models often results in high computational costs and increased inference times. This creates the need to enhance the performance of the classification task. There are two approaches which can be leveraged to improve the efficiency of long document classification. They can be broadly classified as data-based approaches and model-based approaches.

\subsection{Model Based Approaches:}

Handling long document classification efficiently requires balancing model complexity with computational feasibility. Various techniques have been explored to achieve this, including sparse attention mechanisms, quantization, recurrent architectures, and normalization techniques.
\newline\newline
Sparse attention mechanisms enable transformer models to process significantly longer inputs while retaining the advantages of full-attention models (\citet{Pham_2024}). By incorporating global tokens for capturing overall context, local tokens for nearby interactions, and random tokens to enhance global coverage, these mechanisms effectively reduce memory and computation costs from quadratic to linear (\citet{Martins_2020}). This makes them particularly useful for long document classification, where handling extensive input sequences efficiently is crucial.
\newline\newline
Beyond attention mechanisms, reducing the computational demand of models is also essential. One effective approach is quantization, which lowers the precision of a model’s weights to reduce memory usage. For instance, Q8 BERT employs 8-bit weights instead of the standard 32-bit, leveraging techniques such as quantization-aware training (\citet{Zafrir_2019}). This significantly reduces model size while maintaining high accuracy, making it an attractive solution for deploying deep learning models on resource-constrained systems.
\newline\newline
While transformers dominate modern NLP tasks, recurrent architectures like Long Short-Term Memory (LSTM) networks have also been explored for capturing long-term dependencies. LSTMs excel at preserving sequential information, making them well-suited for long document processing (\citet{Teragawa_2021}),(\citet{Putri_2023}). However, their sequential nature limits parallelization and scalability, giving transformer-based models an edge in handling large-scale text data more efficiently.
\newline\newline
To further enhance the stability and efficiency of transformer models, pre-layer normalization is applied. This technique normalizes activations before the attention mechanism, mitigating gradient instability and accelerating convergence (\citet{Beltagy_2020}). By improving training dynamics, pre-layer normalization enhances the robustness of deep transformer-based architectures, making them more suitable for long document classification.
\newline\newline
By combining these techniques sparse attention for efficiency, quantization for reduced computational demand, recurrent mechanisms for sequence retention, and pre-layer normalization for stability modern NLP models can effectively process long documents while optimizing performance and resource utilization (\citet{AlQurishi_2022}).
\subsection{Data Based Approaches:}

Unlike model-based approaches, which focus on improving model architectures and algorithms, data-centric approaches modify the training and testing data that is fed into the model. These methods aim to improve model performance and efficiency by optimizing the data pipeline rather than altering the model itself.
\newline\newline
A notable data-centric technique is Discriminative Active Learning (DAL), an advanced strategy designed to minimize labeling effort while maximizing model performance. DAL selects the most informative instances for labeling by identifying data points that lie near the model's decision boundary, making the labeled data distribution indistinguishable from the unlabeled pool in the learned representation space (\citet{Bamman_2013}). This selective sampling process reduces the number of labeled examples required to achieve high accuracy.
\newline\newline
Another effective data-based strategy addresses the input length limitations of transformer-based models. One common approach involves splitting long documents into smaller, manageable segments, which are processed individually before being aggregated for final classification. Hierarchical models are often used in this context, where local information from document chunks is first processed, and then a higher-level model combines the outputs (\citet{Yang_2016,Khandve_2022}). Hierarchical attention mechanisms further enhance efficiency by selectively focusing on the most relevant parts of the document, avoiding the need to encode the entire document at once.
\newline\newline
In our work, we prioritize a data-centric approach due to its seamless integration with existing models, domain-agnostic applicability, robustness against model-induced biases, and scalability across diverse tasks (\citet{Song_2024,Moro_2023}). Specifically, we focus on minimizing the amount of contextual information supplied to the model during both training and inference, thereby improving computational efficiency (\citet{He_2019,Liu_2018b,Tay_2021}). This is achieved by selectively curating and streamlining the input data, rather than altering the model's architecture. Such an approach enables greater adaptability across various domains and model types without the overhead of architectural modifications (\citet{Li_2018,Prabhu_2021,Sun_2020}).

\section{Methodology}
The datasets utilized in our experiments are sourced from L3Cube’s IndicNews corpus, a multilingual text classification dataset curated for Indian regional languages. The MahaNews corresponds to the Marathi subset of the IndicNews dataset. The corpus covers news headlines and articles in 11 prominent Indic languages, with each language dataset encompassing 10 or more news categories. We have made use of the Long Document Classification (LDC) dataset which consists of full articles with their categories (\citet{Mittal_2024,Mirashi_2024}).
\newline
Our methodology focuses on optimizing input size while preserving classification performance using the Marathi LDC dataset, which consists of full-length articles in Marathi (\citet{Jain_2020}). We begin by tokenizing each article into individual sentences, followed by computing the TF-IDF score for each sentence. The sentences are then ranked based on their scores, and the context is reduced by selecting top-ranked sentences. To achieve this, we explore various sentence selection strategies. Instead of using the entire article, the selected sentences are fed into the MahaBERT model for classification.
\subsection{Training and Testing}

We aim to improve classification efficiency by reducing the amount of input text during both training and inference, while maintaining comparable performance to the full-document setup. When trained and evaluated on the full LDC dataset, the MahaBERT model, fine-tuned on L3Cube-MahaCorpus and other public Marathi datasets, achieved an accuracy of 94.706\%. \newline
Our goal is to approach this accuracy using reduced-context inputs, which in turn lowers computational costs and training and testing time. The Marathi LDC dataset includes 20425 samples for training and 2550 samples for testing. Additionally, the dataset includes 2,548 validation samples that help enhance the accuracy of the trained model.
\newline\newline
\textbf{Sentence Selection Techniques}
\newline\newline
To achieve reduced-context classification, we evaluated several sentence selection strategies, ranging from simple selection methods to a novel TF-IDF-based method. These approaches aim to retain the most informative parts of each document which are used to train the model and subsequently test it.

\begin{itemize}
    \item \textbf{First Few Sentences Selection:} 
    In this method, only a specified percentage of the initial sentences from each article is selected, which are used to train and test the model. This leverages the observation that the opening sentences often contain summaries or key contextual information critical for classification.

    \item \textbf{Last Few Sentences Selection:} 
    Conversely, this method selects only the last portion of sentences from each article. The rationale is that concluding sentences often include detailed analysis or summaries, which may also be useful for accurate classification.

    \item \textbf{Random Sentences Selection:} 
    Here, sentences are randomly selected from across the article. While this approach is computationally efficient and allows for diverse content selection, it is unreliable, as critical information may be excluded, leading to inconsistent classification performance across samples.
\end{itemize}
While these methods are straightforward and easy to implement, they can fail to consistently capture the document’s most relevant content, as important sentences may appear in various parts of the text.

\begin{figure*}[]
    \includegraphics[width=\textwidth]{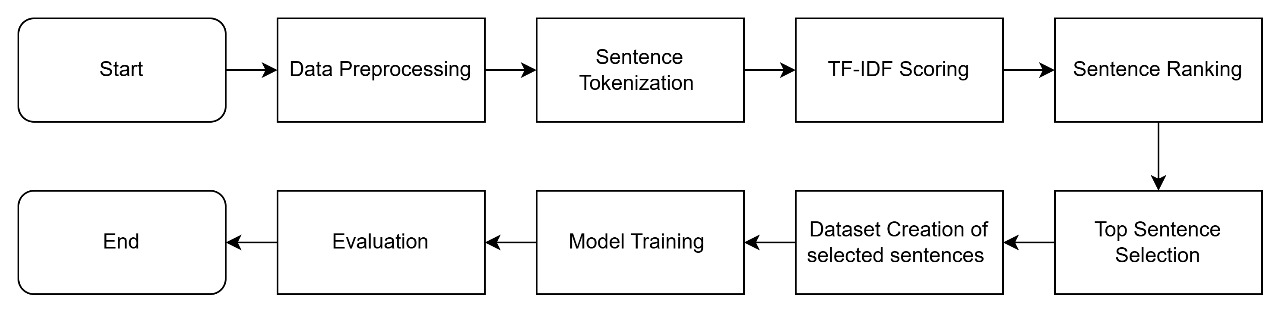}
    \caption{Workflow of ranked approach for sentence selection --- The diagram illustrates a ranked sentence selection workflow starting from raw text input. Data is preprocessed and split into sentences, which are scored using TF-IDF. Top-ranked sentences are selected to create a training dataset. This dataset is used for model training and evaluation before concluding the process.
}
    \label{fig:process_diagram}
\end{figure*}

\subsection{TF-IDF-Based Ranking and Selection }

Random selection of sentences, though computationally efficient, does not account for the relative importance of sentences within a document, which depends on factors such as distinctiveness and semantic relevance. This leads to inconsistent and unreliable classification accuracy, especially in long document classification tasks.
\newline
To address these limitations, we propose a novel sentence selection technique based on TF-IDF scores, which ranks sentences by their informative value. This method significantly reduces inference and training time while maintaining high classification accuracy, offering an effective and scalable solution for long document classification.
\newline\newline
\textbf{General Flow of the Method }
\newline\newline
The sentences of each article are preserved in their original order and tokenized using a language-specific tokenizer, such as the Indic NLP tokenizer for the IndicNews dataset.
\newline
For TF-IDF score calculation, each sentence is treated as an individual document in the context of determining Term Frequency (TF) and Inverse Document Frequency (IDF). This approach identifies terms that occur frequently within a sentence but are rare across others in the same article, thereby quantifying the importance of each term.
\newline
After computing TF-IDF scores for all distinct terms in the article, each sentence receives a cumulative TF-IDF score, aggregated from its constituent terms. Sentences are then ranked based on these cumulative scores, resulting in an ordered array where the most informative sentences appear at the top.
\newline\newline
\textbf{Score Computation}
\newline\newline
The score of a sentence Si can be calculated as the sum of the TF-IDF scores of all terms $t_j$ within the sentence.
\newline
Formally, the score Score($S_i$) is defined as:
\newline\newline
\[
\text{Score}(S_i) = \sum_{t_j \in S_i} \text{TF-IDF}(t_i)
\]

Where,
\[
\text{TF-IDF}(t_i) = \text{TF}(t_i) \cdot \text{IDF}(t_i)
\]

\subsection*{Definitions}

\begin{figure*}
  \centering
  \includegraphics[width=\textwidth]{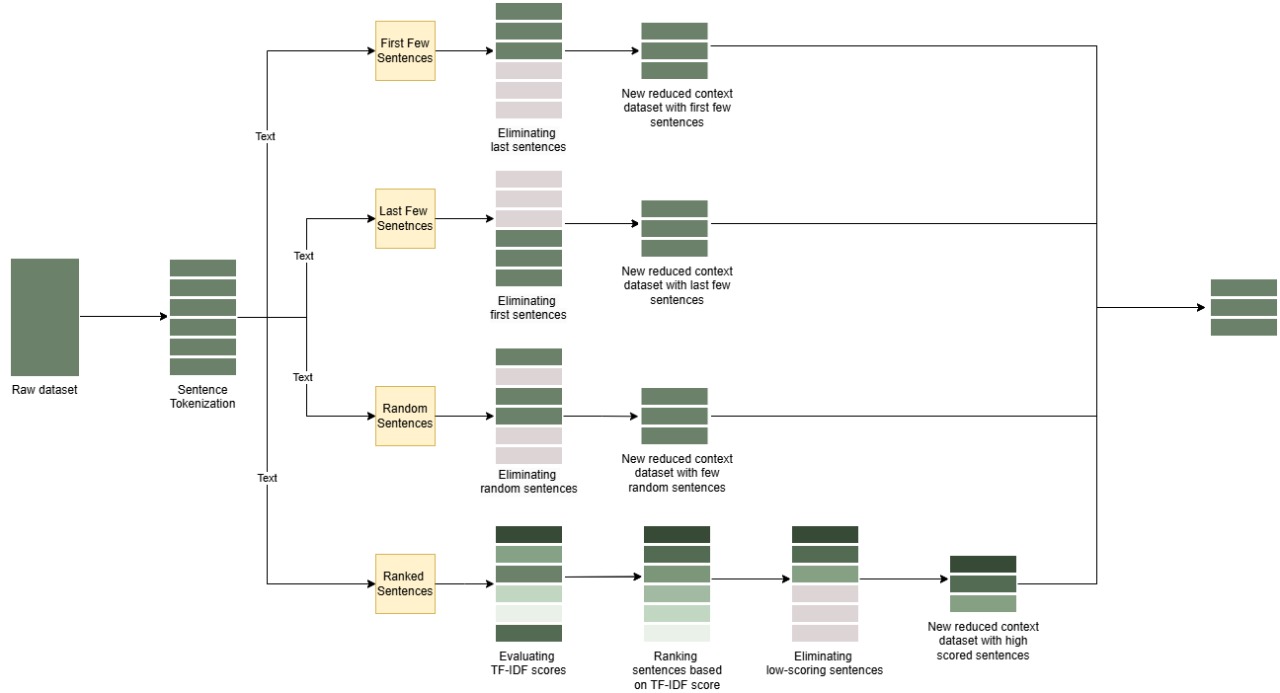}
  \caption{Sentence selection approaches --- The image illustrates various sentence selection approaches used for context reduction. Methods include selecting the first few sentences, last few sentences, random sentences, and ranked sentences. In the ranked approach, sentences are scored using TF-IDF and selected based on relevance. These strategies help reduce input size while preserving meaningful context.
}
  \label{fig:sent_selection}
\end{figure*}

\begin{enumerate}
    \item \textbf{Term Frequency (TF):}
    \[
    \text{TF}(t_j) = \frac{\text{Frequency\ of\ } t_j \text{\ in\ } S_i}{\text{Total\ number\ of\ terms\ in\ } S_i}
    \]

    \item \textbf{Inverse Document Frequency (IDF):}
   \[
\text{IDF}(t_j) = \log\left(\frac{N}{1 + \text{Sentence\ frequency\ of\ } t_j}\right)
\]

\end{enumerate}
where N is the total number of sentences in the article, and the document frequency is the number of sentences containing tj.
\newline
This formula ensures that the importance of each sentence is derived from the significance of its terms within the context of the article.
\newline
The above approaches, to select sentences, for model training are as depicted in Figure \ref{fig:sent_selection}.
\newline\newline
\textbf{Optimal Number Sentence Selection: }
\newline\newline
Selecting the optimal number of sentences involves balancing training efficiency and classification accuracy. Several approaches are considered for determining the most informative subset of sentences:

\begin{itemize}
    \item \textbf{Top-ranked sentence selection:} The highest-ranked sentence, based on its TF-IDF score, is used to evaluate the effectiveness of minimal context in classification.

    \item \textbf{Incremental context expansion:} The top two, three, four, and five ranked sentences are examined to assess the impact of increasing contextual information on classification accuracy and to identify the point of diminishing returns.

    \item \textbf{Percentage-based selection:} Top-ranked sentences are progressively selected in increments of 10\%, aiming to find an optimal balance between efficiency and performance. This method is particularly effective for documents.
\end{itemize}
These approaches help refine sentence selection strategies to enhance both computational efficiency and model performance while minimizing unnecessary information.
\subsection{Length Normalization }

Normalization is a data preprocessing technique that adjusts the values of features or variables to a common scale without distorting differences in the ranges of values. This adjusts the scores to lie within a specific range of value, and ensures that all the features are on the same scale and no single feature dominates because of its scale allowing fair contributions to calculations. 
\newline\newline
In our case, normalization is necessary because we score each sentence based on the sum of its TF-IDF scores and rank them to select the top sentences for training and testing. Without normalization, longer sentences (with more tokens) will naturally have higher TF-IDF sums simply due to having more terms, rather than because they contain more important terms. This creates a bias where longer sentences are ranked higher, even if they do not have proportionally more informative content.
\newline\newline
To ensure fair sentence ranking, different approaches balance sentence length and TF-IDF scores:
\begin{itemize}
    \item \textbf{Length Normalization:} Divides the total TF-IDF score by sentence length to prevent longer sentences from being unfairly ranked higher.


    \item \textbf{Weighted Balancing:} Uses a dynamic weighted formula to balance TF-IDF score and sentence length.
\end{itemize}
By normalizing using sentence length (i.e., dividing the total TF-IDF score by the number of tokens in the sentence), we ensure that the ranking reflects the average importance of terms within a sentence rather than the absolute sum. This helps in fairly comparing sentences of different lengths, preventing a bias toward longer sentences and ensuring that selection is based on term significance rather than sentence size.
\newline
However, after analyzing the selected sentences, it was observed that normalization introduced an inverse bias, causing the algorithm to prioritize shorter sentences. An alternative approach to mitigate this is to incorporate an additional factor that ensures a more balanced and meaningful sentence scoring process.
\newline\newline
\textbf{Balancing Length Factor}
\newline\newline
To achieve a fair ranking, we needed a mechanism that dynamically adjusts the influence of TF-IDF scores and sentence length. This approach creates a flexible ranking mechanism, where the relative importance of each factor can be controlled to ensure an optimal trade-off between uniqueness and context.
\newline
To balance this bias and achieve a trade-off between the two extremes, the following formula was introduced:

\text{Score} = $(\lambda_1 \cdot \text{Normalized\_TF\_IDF}) + (\lambda_2 \cdot \text{length})$
\newline

Where,  $\lambda_1 = 1 - \lambda_2 \quad \text{and} \quad 0 \leq \lambda_1, \lambda_2 \leq 1$
\newline

These are weights to control the relative importance of \textbf{Normalized\_TF\_IDF} and \textbf{length} in the final ranking.
\begin{itemize}
    \item $\lambda_1 > \lambda_2$: Focus on sentences with unique terms (higher TF-IDF score).
    \item $\lambda_2 > \lambda_1$: Prioritize sentences with more context (lengthier ones).
\end{itemize}
This formula effectively balances the biases introduced by normalization and sentence length by distributing the total weight between the two factors. Since $\lambda_1$ = 1 \text{-} $\lambda_2$ , increasing the weight on one factor automatically reduces the influence of the other, ensuring a controlled trade-off. If $\lambda_1$ is higher, the ranking favors sentences with higher TF-IDF scores, emphasizing term uniqueness. Conversely, if $\lambda_2$ is higher, longer sentences with more contextual richness are prioritized. This dynamic weighting mechanism allows for fine-tuning based on the specific needs of the classification task, preventing extreme biases toward either short or long sentences.

\begin{table}[h!]
\resizebox{0.5\textwidth}{!}{ 
\begin{tabular}{|c|c|c|c|c|}
\hline
\textbf{Sentence(S)} & \textbf{First} & \textbf{Last} & \textbf{Random} & \textbf{Ranked} \\ \hline
1 & \textbf{90.70}\% & 81.64\% & 75.76\% & 90.35\% \\ \hline
2 & 93.01\% & 87.60\% & 90.31\% & \textbf{93.17}\% \\ \hline
3 & 93.17\% & 89.76\% & 91.49\% & \textbf{93.64}\% \\ \hline
4 & 92.82\% & 91.09\% & 91.72\% & \textbf{94.00}\% \\ \hline
5 & 93.56\% & 91.64\% & 92.70\% & \textbf{94.19}\% \\ \hline
\end{tabular}
}
\caption{Sentence-wise Accuracy Results — The table shows accuracy across different sentence selection strategies (first, last, random, ranked) for 1 to 5 selected sentences. Results indicate that the ranked approach performs best, followed by first, random, and last, highlighting the importance of the selection method and sentence count on model performance.}
\label{tab:sentence_comparison}
\end{table}
\begin{figure}
    \includegraphics[width=0.5\textwidth]{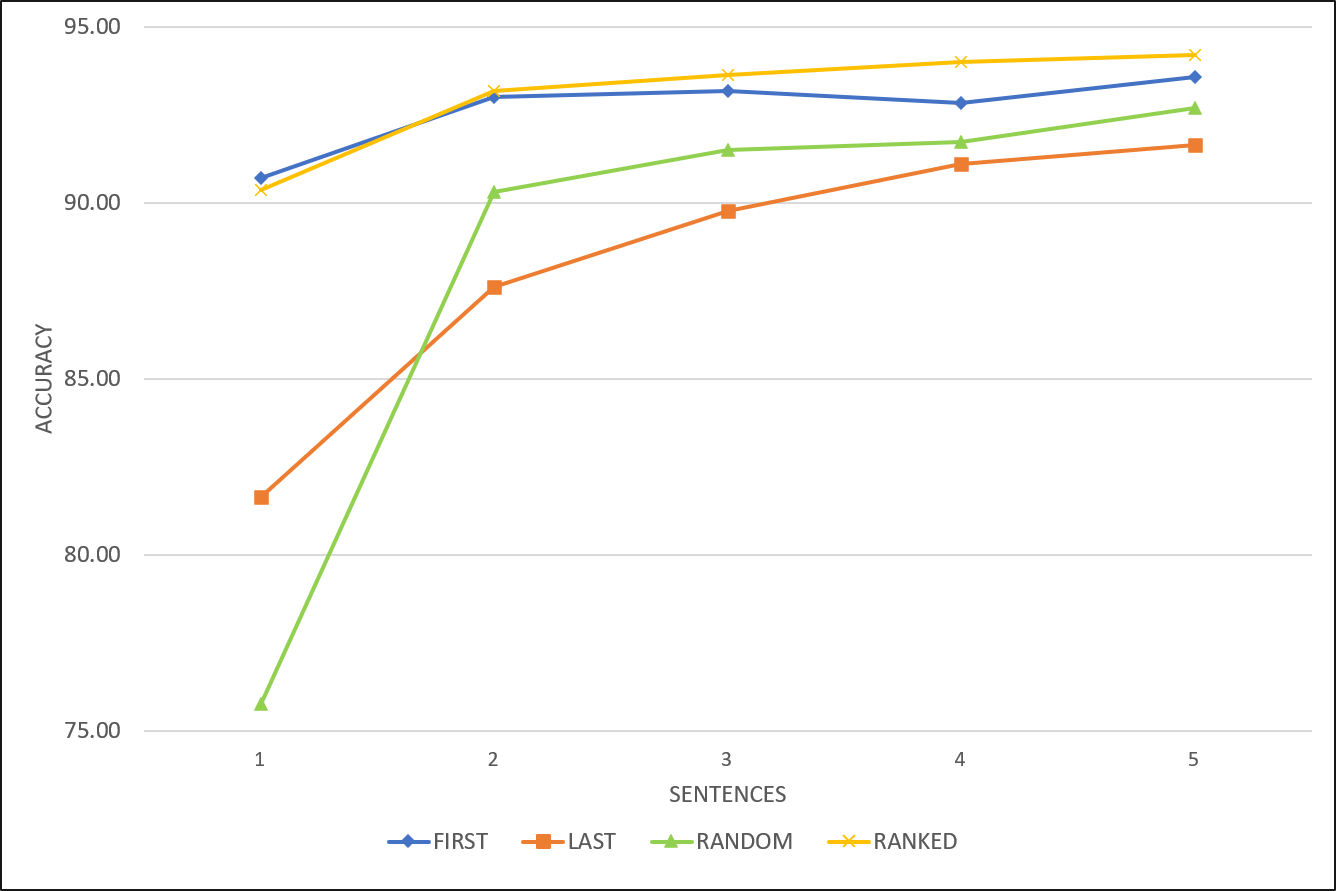}
    \caption{Sentence-wise accuracy graph --- The graph visualizes sentence-wise accuracy for different selection methods: first, last, random, and ranked. It plots accuracy against the number of selected sentences (1 to 5). The graph shows that ranked > first > random > last.}
    \label{fig:graph1}
\end{figure}
\section{Results and Discussion}
\subsection{Number of Sentences Approach}

The results, as presented in Table \ref{tab:sentence_comparison} and Figure \ref{fig:graph1}, show the accuracy corresponding to the selection of the first, last, random, and ranked top 1, 2, 3, 4, and 5 sentences per document. An evident trend is that the accuracy improves as the number of selected sentences increases, reflecting the added informational value of including more content. This improvement is consistently observed across the first, last, and random selection methods, though it peaks at three sentences in the ranked method. Specifically, the performance of selection strategies follows the order: ranked > first > random > last.

By selecting a fixed number of top-ranked sentences, irrespective of document length, the inference time is significantly reduced and remains nearly constant across documents. This approach achieves an accuracy of \textbf{94.19\%} with just 5 sentences, representing a drop of hardly \textbf{0.544\%} from the full-context baseline accuracy of 94.706\%. This demonstrates that substantial reductions in input length can be achieved with minimal impact on accuracy. 

While fixed-length top-ranked sentence selection yields strong results, it treats all high-scoring sentences equally, regardless of their length or proportional informativeness. To further refine this approach, we introduce a normalization strategy that combines normalized TF-IDF scores with sentence length to better capture both relevance and informational content.
\newline \newline
\textbf{Normalization Results}
\begin{table}
\raggedright 
\resizebox{0.5\textwidth}{!}{ 
\begin{tabular}{|c|c|c|c|c|}
\hline
\textbf{Sentence(S)} & \textbf{0.2 ($\lambda_2$)} & \textbf{0.5 ($\lambda_2$)} & \textbf{0.7 ($\lambda_2$)} & \textbf{1.0 ($\lambda_2$)} \\ \hline
1 & 89.11\% & 88.94\% & 89.10\% & \textbf{89.26}\% \\ \hline
2 & \textbf{92.82}\% & 91.86\% & 92.73\% & 92.23\% \\ \hline
3 & 93.79\% & 93.81\% & 93.78\% & \textbf{93.82}\% \\ \hline
4 & 93.95\% & 93.36\% & \textbf{94.07}\% & 93.59\% \\ \hline
5 & 93.47\% & 93.56\% & \textbf{93.67}\% & 93.32\% \\ \hline
\end{tabular}
}
    \caption{Normalized sentence-wise accuracy results --- The table presents normalized sentence-wise accuracy results for ranked sentence selection method. It shows results across different values of $\lambda_2$ ranging from 0.2 to 1.0. $\lambda_2 = 0.7$ provides optimal performance for different sentence counts.}
    \label{fig:normalize_results}
\end{table}

\begin{figure}
    \raggedright 
    \includegraphics[width=0.5\textwidth]{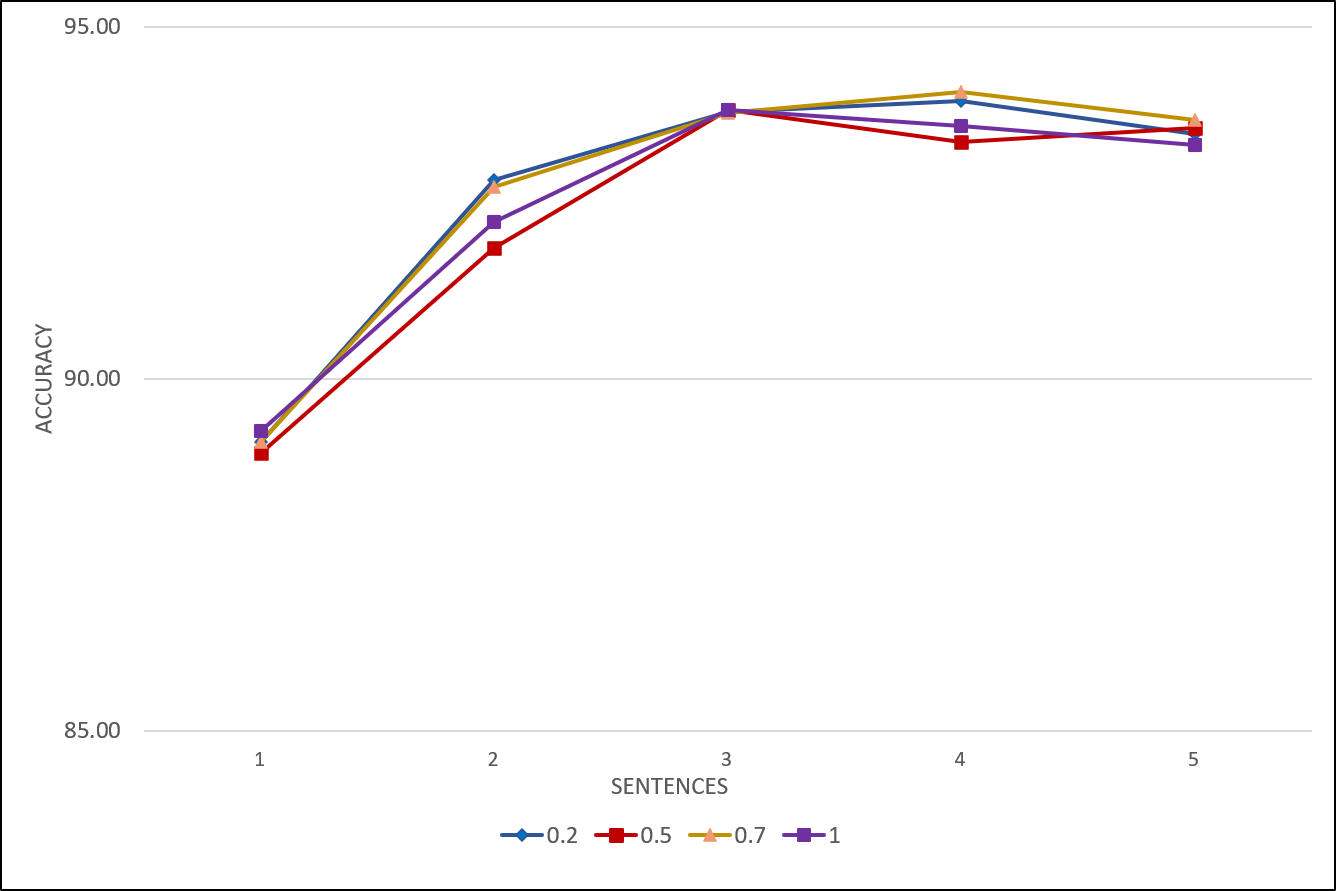}
    \caption{Normalized sentence-wise accuracy graph --- The graph shows normalized sentence-wise accuracy for the ranked selection method. The x-axis represents the number of selected sentences, while the y-axis shows the corresponding accuracy. Each line corresponds to a different $\lambda_2$.}
    \label{fig:graph2}
\end{figure}

Table \ref{fig:normalize_results} shows the impact of combining normalized TF-IDF scores with sentence length using various weightings ($\lambda_2$ = 0.2, 0.5, 0.7, 1.0) across 1 to 5 selected sentences. Accuracy peaks at $\lambda_2$ = 0.7, indicating that longer sentences are more informative in minimal contexts.
As more sentences are included, TF-IDF gains importance. The best results occur at $\lambda_2$ = 0.2 and 0.7, balancing relevance and length. The highest accuracy, \textbf{94.07\%}, is achieved with 4 sentences at $\lambda_2$ = 0.7.
Accuracy stabilizes with higher sentence counts, showing diminishing returns from further tuning as the model becomes less sensitive to weighting changes.

\begin{table}
\raggedright 
\resizebox{0.5\textwidth}{!}{
\begin{tabular}{|c|c|c|}
\hline
\textbf{Sentence(S)} & \textbf{Ranked} & \textbf{Ranked Normalized} \\ \hline
1 & 90.35\% & 89.11\% \\ \hline
2 & 93.17\% & 92.82\% \\ \hline
3 & 93.64\% & 93.82\% \\ \hline
4 & 94.00\% & 94.07\% \\ \hline
5 & 94.19\% & 93.67\% \\ \hline
\end{tabular}
}
\caption{Comparison of Ranked and Ranked-Normalized Results — The table compares the accuracy of sentence selection using ranked and ranked-normalized methods. It shows that normalization has little impact when the number of selected sentences is low.}
\label{tab:ranked_comparison}
\end{table}

Table \ref{tab:ranked_comparison} compares the classification performance of the basic ranked sentence selection approach with the normalized ranking strategy, where normalized TF-IDF scores are combined with sentence length using optimal weighting parameters ($\lambda_1$ and $\lambda_2$) for each case. The results show no significant improvement with normalization, suggesting that this approach has minimal impact when only a few sentences are selected.
\newline
\subsection{Data Percentage Approach}
\renewcommand{\arraystretch}{1.5} 
\begin{table}
\fontsize{15}{20}\selectfont
\raggedright 
\resizebox{0.5\textwidth}{!}{ 
\begin{tabular}{|c|c|c|c|c|c|}
\hline
\textbf{Percentage} & \textbf{First} & \textbf{Last} & \textbf{Random} & \textbf{Ranked} & \textbf{Ranked Normalized} \\ \hline
10\% & 90.74\% & 71.76\% & 86.00\% & \textbf{91.80}\% & 91.14\% \\ \hline
20\% & 93.25\% & 87.88\% & 90.31\% & 93.41\% & \textbf{93.64}\% \\ \hline
30\% & 93.19\% & 90.31\% & 92.22\% & 93.29\% & \textbf{93.80}\% \\ \hline 
40\% & 93.58\% & 91.96\% & 92.82\% & 93.98\% & \textbf{94.39}\% \\ \hline 
50\% & 94.19\% & 92.98\% & 93.33\% & \textbf{94.51}\% & 94.04\% \\ \hline
60\% & \textbf{94.31}\% & 92.86\% & 93.05\% & \textbf{94.31}\% & 93.60\% \\ \hline
70\% & 94.62\% & 94.19\% & 94.31\% & 93.92\% & \textbf{94.47}\% \\ \hline
80\% & \textbf{94.43}\% & 93.45\% & 94.23\% & 94.15\% & 94.15\% \\ \hline
90\% & 94.50\% & 93.56\% & 94.03\% & 94.11\% & \textbf{94.90}\% \\ \hline
100\% & 94.35\% & 94.35\% & 94.11\% & 94.63\% & \textbf{94.78}\% \\ \hline
\end{tabular}
}
\caption{Percentage-wise accuracy results --- The table presents percentage-wise accuracy results for sentence selection approaches at coverage levels from 10$\%$ to 100$\%$. It compares First, Last, Random, Ranked, and Ranked Normalized methods.}
\label{tab:percentage_comparison}
\end{table}
   
\begin{figure}
    \raggedright 
    \includegraphics[width=0.5\textwidth]{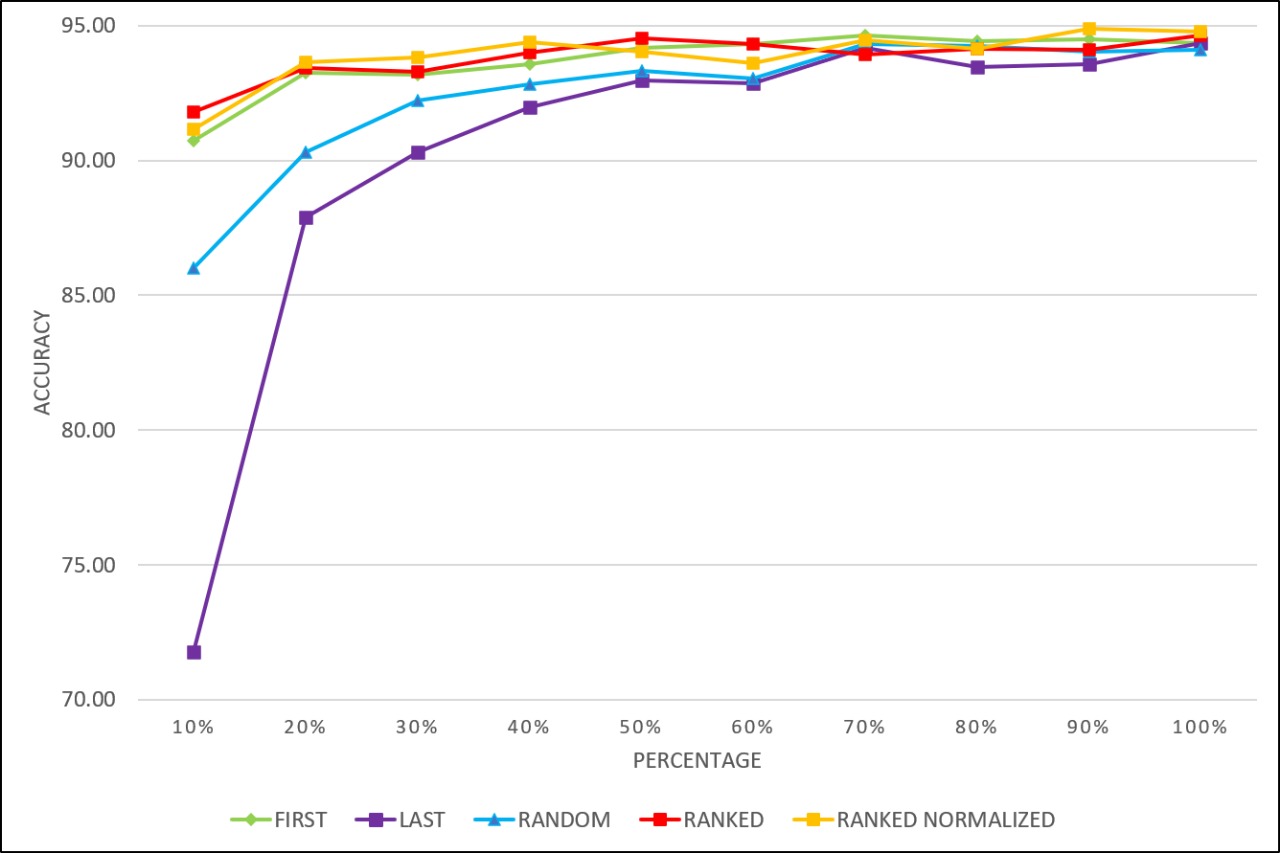}
    \caption{Percentage-wise accuracy graph --- The graph visualizes percentage-wise accuracy for different sentence selection methods as sentence coverage increases from 10$\%$ to 100$\%$. The x-axis represents the percentage of selected sentences, while the y-axis shows accuracy. Each line corresponds to a method: First, Last, Random, Ranked, and Ranked Normalized. The graph highlights how accuracy improves with more context and which methods are most effective.
}
    \label{fig:graph3}
\end{figure}
Table \ref{tab:percentage_comparison} and Figure \ref{fig:graph3} illustrate the accuracy achieved by selecting first, last, random, and ranked percentages of sentences from documents. Using the full-length documents for training and testing yields an accuracy of 94.706\%, which serves as the baseline for comparison. Importantly, by reducing the context to just 40 to 50 percent of the original document, we are still able to achieve an impressive accuracy of \textbf{94.39\%}, remarkably close to the base accuracy of 94.706\%, demonstrating that its performance is competitive with approaches that use the full document context.
\newline\newline
With reduced context sizes, the Ranked Selection method consistently outperforms other techniques, such as First, Last, and Random selection. As the context size increases, the performance of all methods converges, yielding similar results. This convergence indicates that the Ranked Selection method is particularly effective in enhancing accuracy when operating with smaller context windows.  In this setting, normalization shows a positive impact, enhancing performance in most cases.
\subsection{Inference Time}

The goal of context reduction is to minimize inference time without sacrificing classification accuracy. Our approach uses TF-IDF to trim context effectively, aligning the sequence length with each document’s actual content rather than relying on fixed limits (e.g., 512 tokens in BERT). Fixed lengths with padding or truncation undermine efficiency gains, while dynamic adjustment ensures both accurate and faster inference.
The graphs in Figure \ref{fig:graph4} and \ref{fig:graph5} below depict the change in testing times while selecting varying contexts.
\begin{figure}
    \raggedright 
    \includegraphics[width=0.5\textwidth]{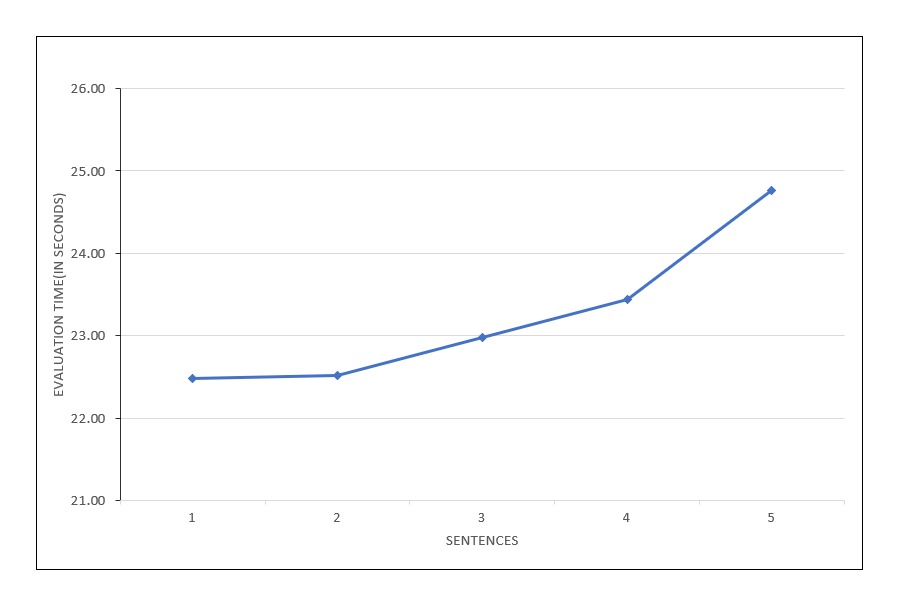}
    \caption{Evaluation time graph for sentence-wise selection --- The graph represents the relationship between evaluation time (in seconds) and the number of sentences considered during sentence-wise selection. The x-axis denotes the number of sentences,  while the y-axis shows the corresponding evaluation time required. The graph typically highlights a trend where evaluation time increases as the number of sentences grows}
    \label{fig:graph4}
\end{figure}
\newline
Figure \ref{fig:graph4} shows a positive correlation between the number of sentences and evaluation time. While the increase is modest from 1 to 3 sentences, it becomes more pronounced from sentence 4 onward, indicating that evaluation time grows increasingly with higher sentence counts.
\begin{figure}
    \raggedright 
    \includegraphics[width=0.5\textwidth]{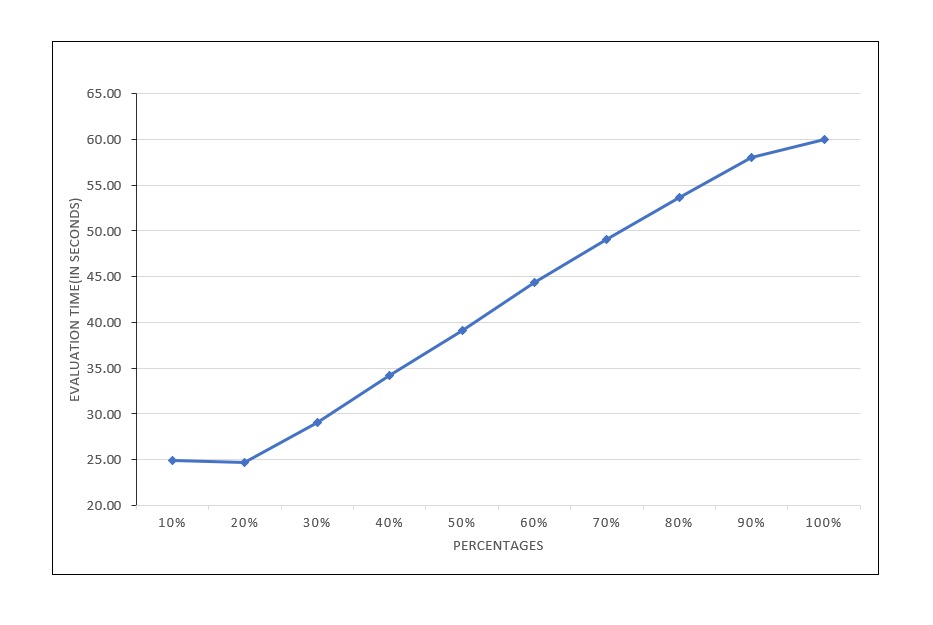}
    \caption{Evaluation time graph for percentage-wise selection --- The graph illustrates the relationship between evaluation time (in seconds) and the percentage of sentences selected during percentage-wise selection. The graph demonstrates a trend where evaluation time changes based on the percentage selected.}
    \label{fig:graph5}
\end{figure}
\newline
Figure \ref{fig:graph5} illustrates the relationship between the percentage of content evaluated and the corresponding evaluation time. Evaluation time remains relatively stable from 10\% to 20\%, then increases sharply from 30\% onward, peaking at 100\%. This indicates a strong positive—and increasingly non-linear—correlation, especially at the extremes, likely due to higher computational load.

Notably, at just 40\% of the original document context, our method achieves 94.39\% accuracy—only 0.33\% below the full-context baseline—while reducing inference latency by 43\%. This demonstrates an efficient trade-off between speed and performance, making the approach well-suited for real-world applications demanding both accuracy and scalability.

\section{Conclusion}

We introduced an efficient approach to long document classification by leveraging sentence selection techniques that reduce input size while maintaining accuracy comparable to full-context models. Traditional transformer-based models struggle with long texts due to input length constraints and computational costs. Our method addresses these limitations through strategies such as first/last sentence selection, random sampling, and TF-IDF-based ranking.

Experiments on the Marathi Long Document Classification (LDC) dataset from L3Cube's IndicNews collection show that our approach significantly reduces computational costs without sacrificing performance. TF-IDF-based ranking proved especially effective in identifying informative sentences for classification.

We also explored the trade-off between input size and accuracy, demonstrating that selecting a proportion of high-ranking sentences—rather than a fixed number—yields a better balance between efficiency and performance. Incorporating normalization into ranking further improves accuracy, even with reduced input size.

In summary, our method offers a scalable, resource-efficient solution for long document classification. Future work may involve domain-specific sentence selection or hybrid models to further refine input representation and enhance performance across diverse NLP tasks.

\section*{Acknowledgement}
This work was undertaken with the mentorship of L3Cube, Pune. We sincerely appreciate the invaluable guidance and consistent encouragement provided by our mentor during this endeavor.








\bibliography{main}

\begin{thebibliography}{32}
\providecommand{\natexlab}[1]{#1}
\providecommand{\url}[1]{\texttt{#1}}
\expandafter\ifx\csname urlstyle\endcsname\relax
  \providecommand{\doi}[1]{doi: #1}\else
  \providecommand{\doi}{doi: \begingroup \urlstyle{rm}\Url}\fi

\bibitem[Aishwarya et~al.(2023)Aishwarya, Srushti, Purva, Tejas, and Raviraj]{Mirashi_2024}
Mirashi Aishwarya, Sonavane Srushti, Lingayat Purva, Padhiyar Tejas, and Joshi Raviraj.
\newblock L3cube-indicnews: News-based short text and long document classification datasets in indic languages.
\newblock In \emph{Proceedings of the 20th International Conference on Natural Language Processing (ICON)}, pages 442--449, 2023.

\bibitem[Al-Qurishi(2022)]{AlQurishi_2022}
Muhammad Al-Qurishi.
\newblock Recent advances in long documents classification using deep-learning.
\newblock In \emph{Proceedings of the 5th International Conference on Natural Language and Speech Processing (ICNLSP 2022)}, 2022.
\newblock \doi{10.18653/v1/2022.icnlsp-1.12}.

\bibitem[Bamman and Smith(2013)]{Bamman_2013}
David Bamman and Noah Smith.
\newblock New alignment methods for discriminative book summarization, 2013.

\bibitem[Beltagy et~al.(2020)Beltagy, Peters, and Cohan]{Beltagy_2020}
Iz~Beltagy, Matthew~E. Peters, and Arman Cohan.
\newblock Longformer: The long-document transformer, 2020.

\bibitem[Das and Chakraborty(2020)]{Das_2020}
Bijoyan Das and Sarit Chakraborty.
\newblock An improved text sentiment classification model using tf-idf and next word negation, 2020.

\bibitem[Das et~al.(2023)Das, Selvakumar, and Alphonse]{Das_2023}
Mamata Das, K.~Selvakumar, and P.J.A. Alphonse.
\newblock A comparative study on tf-idf feature weighting method and its analysis using unstructured dataset, 2023.

\bibitem[Devlin et~al.(2018)Devlin, Chang, Lee, and Toutanova]{Devlin_2018}
J.~Devlin, M.~W. Chang, K.~Lee, and K.~Toutanova.
\newblock Bert: Pre-training of deep bidirectional transformers for language understanding, 2018.

\bibitem[He(2019)]{He_2019}
Jun He.
\newblock Long document classification from local word glimpses via recurrent attention learning.
\newblock \emph{IEEE Access}, 2019.
\newblock \doi{10.1109/ACCESS.2019.2907992}.

\bibitem[Jain et~al.(2020)Jain, Deshpande, Shridhar, et~al.]{Jain_2020}
Kushal Jain, Adwait Deshpande, Kumar Shridhar, et~al.
\newblock Indic-transformers: An analysis of transformer language models for indian languages, 2020.

\bibitem[Joshi(2022)]{joshi2022l3cubeMahaCorpus}
Raviraj Joshi.
\newblock L3cube-mahacorpus and mahabert: Marathi monolingual corpus, marathi bert language models, and resources.
\newblock In \emph{Proceedings of the WILDRE-6 Workshop within the 13th Language Resources and Evaluation Conference}, pages 97--101, 2022.

\bibitem[Khandve et~al.(2022)Khandve, Wagh, Wani, Joshi, and Joshi]{Khandve_2022}
Snehal~Ishwar Khandve, Vedangi~Kishor Wagh, Apurva~Dinesh Wani, Isha~Mandar Joshi, and Raviraj~Bhuminand Joshi.
\newblock Hierarchical neural network approaches for long document classification.
\newblock In \emph{Proceedings of the 2022 14th International Conference on Machine Learning and Computing}, pages 115--119, 2022.

\bibitem[Kim and Gil(2019)]{Kim_2019}
Sang-Woon Kim and Joon-Min Gil.
\newblock Research paper classification systems based on tf-idf and lda schemes.
\newblock \emph{Human-centric Computing and Information Sciences}, 2019.
\newblock \doi{10.1186/s13673-019-0192-7}.

\bibitem[Li et~al.(2018)Li, Cheng, and Wang]{Li_2018}
Chao Li, Yanfen Cheng, and Hongxia Wang.
\newblock A novel document classification algorithm based on statistical features and attention mechanism, 2018.

\bibitem[Liu et~al.(2018{\natexlab{a}})Liu, Sheng, Wei, and Yang]{Liu_2018}
C.~z. Liu, Y.~x. Sheng, Z.~q. Wei, and Y.~Q. Yang.
\newblock Research of text classification based on improved tfi-df algorithm.
\newblock In \emph{2018 IEEE International Conference of Intelligent Robotic and Control Engineering (IRCE)}, 2018{\natexlab{a}}.
\newblock \doi{10.1109/IRCE.2018.8492945}.

\bibitem[Liu et~al.(2018{\natexlab{b}})Liu, Liu, Cong, Zhao, Ji, and He]{Liu_2018b}
L.~Liu, K.~Liu, Z.~Cong, J.~Zhao, Y.~Ji, and J.~He.
\newblock Long length document classification by local convolutional feature aggregation.
\newblock \emph{Algorithms}, 2018{\natexlab{b}}.
\newblock \doi{10.3390/a11080109}.

\bibitem[Martins et~al.(2020)Martins, Farinhas, Treviso, et~al.]{Martins_2020}
André F.~T. Martins, António Farinhas, Marcos Treviso, et~al.
\newblock Sparse and continuous attention mechanisms, 2020.

\bibitem[Minaee et~al.(2021)Minaee, Kalchbrenner, Cambria, et~al.]{Minaee_2021}
S.~Minaee, N.~Kalchbrenner, E.~Cambria, et~al.
\newblock Deep learning based text classification: A comprehensive review, 2021.

\bibitem[Mittal et~al.(2023)Mittal, Magdum, Hiwarkhedkar, Dhekane, and Joshi]{Mittal_2024}
Saloni Mittal, Vidula Magdum, Sharayu Hiwarkhedkar, Omkar Dhekane, and Raviraj Joshi.
\newblock L3cube-mahanews: News-based short text and long document classification datasets in marathi.
\newblock In \emph{International Conference on Speech and Language Technologies for Low-resource Languages}, pages 52--63. Springer, 2023.

\bibitem[Moro(2023)]{Moro_2023}
G.~Moro.
\newblock Efficient memory-enhanced transformer for long-document summarization in low-resource regimes.
\newblock \emph{Sensors}, 2023.
\newblock \doi{10.3390/s23073542}.

\bibitem[Park et~al.(2022)Park, Vyas, and Shah]{Park_2022}
Hyunji Park, Yogarshi Vyas, and Kashif Shah.
\newblock Efficient classification of long documents using transformers.
\newblock In \emph{Proceedings of the 60th Annual Meeting of the Association for Computational Linguistics (Volume 2: Short Papers)}, 2022.
\newblock \doi{10.18653/v1/2022.acl-short.79}.

\bibitem[Pham and The(2024)]{Pham_2024}
Linh~Manh Pham and Hoang~Cao The.
\newblock Lnlf-bert: Transformer for long document classification with multiple attention levels.
\newblock \emph{IEEE Access}, 2024.
\newblock \doi{10.1109/ACCESS.2024.3492102}.

\bibitem[Prabhu et~al.(2021)Prabhu, Mohamed, and Misra]{Prabhu_2021}
Sumanth Prabhu, Moosa Mohamed, and Hemant Misra.
\newblock Multi-class text classification using bert-based active learning, 2021.

\bibitem[Putri and Setiawan(2023)]{Putri_2023}
Bella~Adriani Putri and Erwin~Budi Setiawan.
\newblock Topic classification using the long short-term memory (lstm) method with fasttext feature expansion on twitter.
\newblock 2023.
\newblock \doi{10.1109/ICoDSA58501.2023.10277033}.

\bibitem[Qaiser and Ali(2018)]{Qaiser_2018}
Shahzad Qaiser and Ramsha Ali.
\newblock Text mining: Use of tf-idf to examine the relevance of words to documents.
\newblock \emph{IJCA}, 2018.
\newblock \doi{10.5120/ijca2018917395}.

\bibitem[Song(2024)]{Song_2024}
B.~Song.
\newblock State space models based efficient long documents classification.
\newblock \emph{Journal of Intelligent Learning Systems and Applications}, 2024.
\newblock \doi{10.4236/jilsa.2024.163009}.

\bibitem[Sun et~al.(2020)Sun, Qiu, Xu, et~al.]{Sun_2020}
Chi Sun, Xipeng Qiu, Yige Xu, et~al.
\newblock How to fine-tune bert for text classification?, 2020.

\bibitem[Tay et~al.(2021)Tay, Dehghani, Abnar, et~al.]{Tay_2021}
Y.~Tay, M.~Dehghani, S.~Abnar, et~al.
\newblock Long range arena: A benchmark for efficient transformers, 2021.

\bibitem[Teragawa et~al.(2021)Teragawa, Wang, and Ma]{Teragawa_2021}
Shoryu Teragawa, Lei Wang, and Ruixin Ma.
\newblock A deep neural network approach using convolutional network and long short term memory for text sentiment classification.
\newblock In \emph{2021 IEEE 24th International Conference on Computer Supported Cooperative Work in Design (CSCWD)}, 2021.
\newblock \doi{10.1109/CSCWD49262.2021.9437871}.

\bibitem[Wagh et~al.(2021)Wagh, Khandve, Joshi, Wani, Kale, and Joshi]{Wagh_2021}
Vedangi Wagh, Snehal Khandve, Isha Joshi, Apurva Wani, Geetanjali Kale, and Raviraj Joshi.
\newblock Comparative study of long document classification.
\newblock In \emph{TENCON 2021-2021 IEEE Region 10 Conference (TENCON)}, pages 732--737. IEEE, 2021.

\bibitem[Yang et~al.(2016)Yang, Yang, Dyer, et~al.]{Yang_2016}
Zichao Yang, Diyi Yang, Chris Dyer, et~al.
\newblock Hierarchical attention networks for document classification.
\newblock In \emph{Proceedings of the 2016 Conference of the North {A}merican Chapter of the Association for Computational Linguistics: Human Language Technologies}, 2016.
\newblock \doi{10.18653/v1/N16-1174}.

\bibitem[Zafrir(2019)]{Zafrir_2019}
O.~Zafrir.
\newblock Q8bert: Quantized 8bit bert, 2019.

\bibitem[Zaheer et~al.(2020)Zaheer, Ainslie, Guruganesh, et~al.]{Zaheer_2020}
M.~Zaheer, J.~Ainslie, G.~Guruganesh, et~al.
\newblock Big bird: Transformers for longer sequences.
\newblock In \emph{Proceedings of NeurIPS}, pages 702--709, 2020.
\newblock \doi{10.48550/arXiv.2007.14062}.

\end{thebibliography}
\bibliographystyle{plainnat}

\end{document}